\pdfoutput=1

\documentclass[11pt]{article}

\usepackage[preprint]{acl}

\AtBeginDocument{\nolinenumbers}

\usepackage{times}
\usepackage{latexsym}

\usepackage[T1]{fontenc}
\usepackage[utf8]{inputenc}
\usepackage{microtype}

\usepackage{amsmath}
\usepackage{amssymb}
\usepackage{amsbsy}
\usepackage{amsthm}
\usepackage{tcolorbox}
\theoremstyle{definition}
\newtheorem{proposition}{Proposition}

\usepackage{graphicx}
\usepackage{float}

\usepackage{enumitem}

\usepackage{array}
\newcolumntype{L}[1]{>{\raggedright\arraybackslash}p{#1}}
\newcolumntype{C}[1]{>{\centering\arraybackslash}p{#1}}
\newcolumntype{R}[1]{>{\raggedleft\arraybackslash}p{#1}}

\usepackage[edges]{forest}
\usepackage{tikz}

\usepackage{booktabs}
\usepackage{multirow}
\usepackage{pifont}
\usepackage{fontawesome5}
\usepackage{colortbl}

\definecolor{shadecolor}{gray}{0.94}
\AtBeginEnvironment{proposition}{%
  \begin{tcolorbox}[
    colback=shadecolor,
    colframe=shadecolor,
    boxrule=0pt,
    arc=0pt,
    left=4pt,
    right=4pt,
    top=3pt,
    bottom=3pt,
    before skip=6pt,
    after skip=6pt,
  ]%
}
\AtEndEnvironment{proposition}{\end{tcolorbox}}

\definecolor{ptcol}{HTML}{1F4E79}    
\definecolor{sftcol}{HTML}{2E7D32}   
\definecolor{prefcol}{HTML}{B5651D}  
\definecolor{gencol}{HTML}{6A4C93}   
\definecolor{rootcol}{HTML}{4A4A4A}  
\newcommand{\tableheadrow}{\rowcolor{rootcol!12}}
\newcommand{\tablestripes}{\rowcolors{2}{rootcol!5}{white}}
\definecolor{objcol}{HTML}{B03A2E}   
\definecolor{polcol}{HTML}{2E7D32}   
\definecolor{updcol}{HTML}{1F5FA8}   
\newcommand{\objJ}{{\color{objcol}\mathcal{J}}}
\newcommand{\policy}{{\color{polcol}\pi_\phi}}
\newcommand{\update}{{\color{updcol}\mathcal{U}}}
\definecolor{findingblue}{HTML}{2F6B7C}
\definecolor{findinggreen}{HTML}{3B7A57}

\renewenvironment{abstract}{%
  \begin{center}
  \begin{tcolorbox}[
    width=0.94\textwidth,
    colback=black!2,
    colframe=black!25,
    boxrule=0.45pt,
    arc=1mm,
    left=14pt,
    right=14pt,
    top=10pt,
    bottom=11pt,
    before skip=6pt,
    after skip=6pt
  ]%
  {\centering\large\bfseries Abstract\par}\vspace{4pt}%
  \normalsize\noindent\ignorespaces
}{%
  \unskip
  \end{tcolorbox}
  \end{center}
}

\newcommand{\insighticon}{%
  \tikz[baseline=-0.55ex,scale=0.16,line width=1.2pt]
    \draw (0,0) circle (1) (0.72,-0.72)--(1.65,-1.65);%
}
\newcommand{\finding}[2]{%
  \begingroup
  \setlength{\fboxsep}{4pt}%
  \noindent\colorbox{#1!6}{%
    \begin{tabular}{@{\color{#1}\vrule width 2pt\hspace{0.6em}}%
                    p{\dimexpr\columnwidth-2\fboxsep-2pt-0.6em\relax}@{}}
      \textcolor{#1!80!black}{\insighticon\hspace{0.45em}\textbf{Finding.}}%
      \hspace{0.25em}#2
    \end{tabular}%
  }\par\nobreak\vspace{4pt}
  \endgroup
}

\newcommand{\keypoint}[2]{%
  \begingroup
  \setlength{\fboxsep}{4pt}%
  \noindent\colorbox{#1!8}{%
    \begin{tabular}{@{\color{#1}\vrule width 2pt\hspace{0.65em}}%
                    p{\dimexpr\columnwidth-2\fboxsep-2pt-0.65em\relax}@{}}
      \textcolor{#1!80!black}{%
        \makebox[1.35em][l]{\raisebox{0.08ex}{\textcircled{\scriptsize\bfseries ?}}}%
        \textbf{#2}}
    \end{tabular}%
  }\par\nobreak\vspace{3pt}
  \endgroup
}

\newcommand{\cmark}{\textcolor{sftcol}{\ding{51}}}
\newcommand{\xmark}{\textcolor{gray!60}{\ding{55}}}

\newcommand{\dataset}{{\cal D}}

\makeatletter
\newcommand{\makesinglecolumntitle}{%
  \par
  \begingroup
    \def\thefootnote{\fnsymbol{footnote}}%
    \@maketitle
    \@thanks
  \endgroup
  \setcounter{footnote}{0}%
  \let\@maketitle\relax
  \gdef\@thanks{}%
  \gdef\@author{}%
  \gdef\@title{}%
  \let\thanks\relax
}
\makeatother

\title{\raisebox{-0.33\height}{\includegraphics[height=1.15cm]{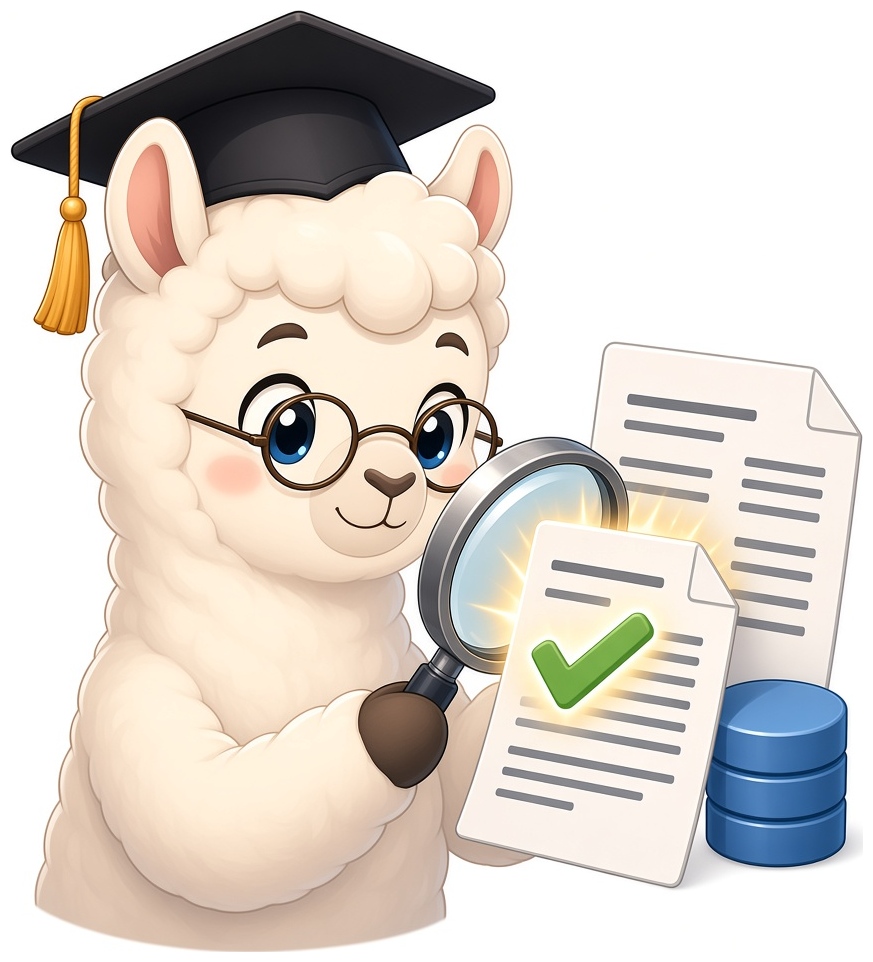}}~Data Strategies for Large Language Models:\\ A Unified Perspective on the Training Lifecycle}

\author{%
  {\bfseries
    Yuanjian Xu$^{1}$, Jianing Hao$^{1}$, Tianze Sun$^{2}$,
    Yongyao Duan$^{3}$, Sam Chen$^{3}$} \\
  {\bfseries
    Jinyi Han$^{4}$, Wanbo Zhang$^{5}$, Yuxuan Sun$^{6}$,
    Guang Zhang\textsuperscript{1,\faEnvelope},
    Zhong Li\textsuperscript{7,\faEnvelope}} \\
  {\normalfont\footnotesize
    $^{1}$The Hong Kong University of Science and Technology (Guangzhou)
    \quad $^{2}$Harbin Institute of Technology} \\
  {\normalfont\footnotesize
    $^{3}$OpenCSG \quad $^{4}$East China Normal University
    \quad $^{5}$Fudan University} \\
  {\normalfont\footnotesize
    $^{6}$University College London
    \quad $^{7}$Microsoft Research Asia
    \quad \mbox{\faEnvelope\ Corresponding authors}}
}

\begin{document}
\onecolumn
\makesinglecolumntitle

\begin{abstract}
Data strategy plays a central role in training large language models (LLMs).
It governs the selection, weighting, and scheduling of training data, thereby
determining the capabilities a model can acquire from a fixed data budget.
We treat high-quality data as having high training-dependent utility, which
depends on the target objective, the model, and its training state rather
than on the text alone. Research in this area now spans pretraining and post-training, but related
ideas use inconsistent terminology and evaluation protocols. This work
reviews data strategies across the training lifecycle and formulates them as
policies that allocate a training budget over candidate examples. We organise
methods by their decision granularity, model dependence, and temporal
adaptivity, then compare their computational requirements and empirical
support. Fine-grained or state-dependent policies offer more control but incur
additional scoring, search, or rollout costs. Evidence for transfer across
model scales and comparable end-to-end cost remains limited. A continuously
updated list of papers, along
with their public implementations, is available at
\url{https://github.com/xuyj233/data_strategy_survey}.
\end{abstract}

\vspace{-0.08in}
\begin{center}
  \includegraphics[width=0.97\textwidth]{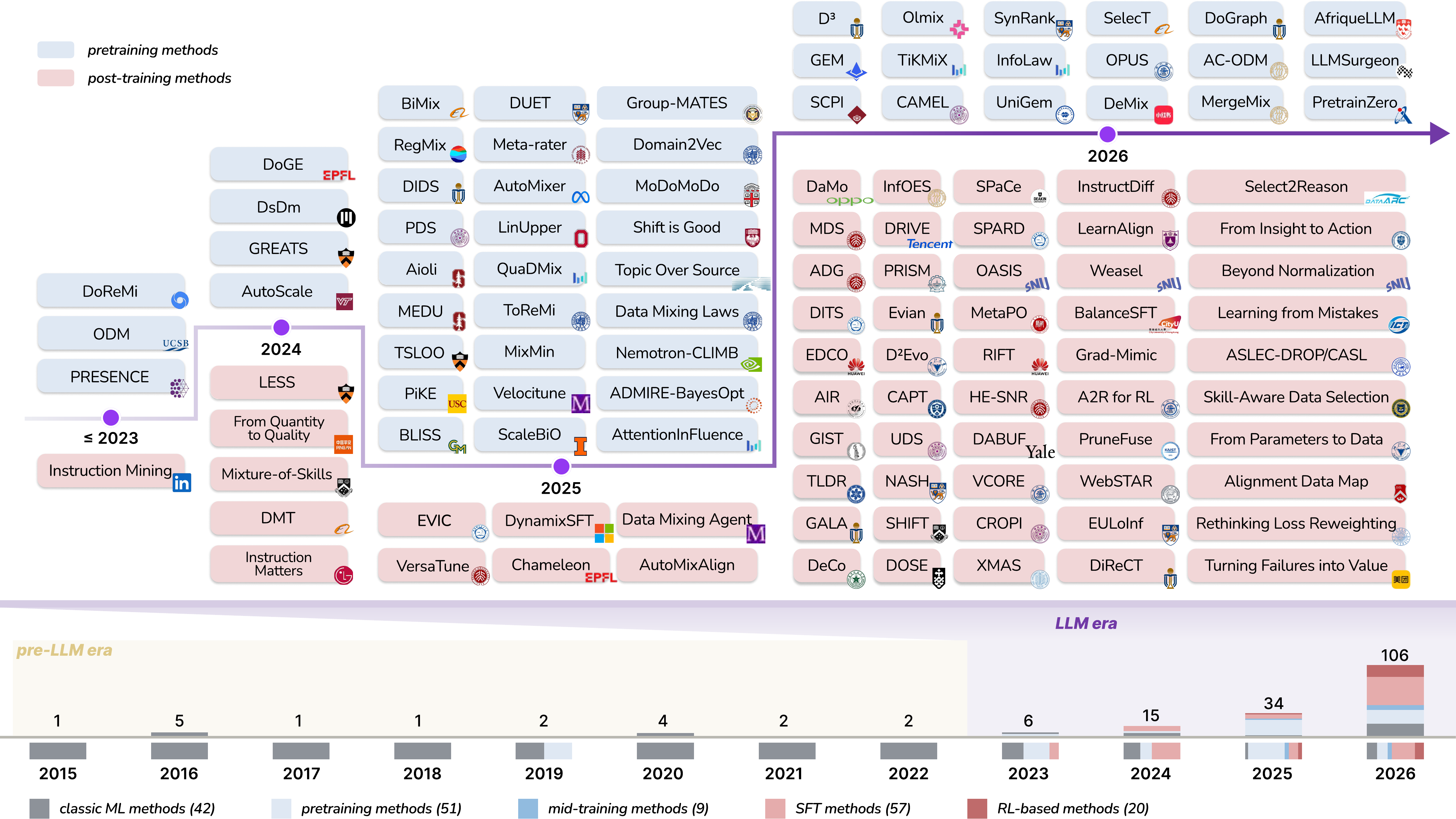}
  \captionsetup{hypcap=false}
  \captionof{figure}{\textbf{Landscape of LLM data strategies.}
  Methods are organised by pretraining and post-training stage.}
  \label{fig:literature-landscape}
\end{center}
\vspace{0.04in}

\section{Introduction}
\label{sec:intro}

Training data is a major determinant of large language model (LLM)
performance~\citep{penedo2023refinedweb,penedo2024fineweb,li2024dclm}.
Scaling laws quantify the effects of model size, compute, and token budget,
whereas data strategy determines which examples fill that
budget~\citep{kaplan2020scaling,hoffmann2022chinchilla}. Data quality and
composition affect both downstream accuracy and training efficiency
\citep{DBLP:conf/nips/Xie0DDLLLL0Y23,DBLP:conf/iclr/LiuZMZDPJL25,DBLP:conf/iclr/YeLSZZQ25}.
Figure~\ref{fig:literature-landscape} traces the expansion from pretraining
mixtures to supervised fine-tuning (SFT), reinforcement learning (RL), and
mid-training.

Across the LLM training lifecycle, data strategies\footnote{Including
selection, reweighting, and curricula.} are shaped by the
distinct constraints of each phase. Pretraining
comprises stable training and annealing, with most work focusing on data
strategies during stable training. Owing to the scale of pretraining corpora,
methods in this phase typically optimise
mixtures at the domain level, commonly referred to as data mixing or
domain reweighting, by assigning one weight to each source or cluster
\citep{DBLP:conf/nips/Xie0DDLLLL0Y23,DBLP:conf/icml/FanPJ24,diao2025nemotronclimb}.
Domain-level optimisation trades sample-level flexibility for tractability,
assuming the corpus admits a useful partition. During annealing, empirical approaches typically advocate
selecting high-quality data \citep{dubey2024llama3,feng2024twophase}. From a
theoretical perspective, \citet{xu2026direct} characterise annealing with a
river-valley model and argue that controlling updates along specific directions
guides the model towards a local minimum.

Post-training data strategy is a multi-granularity control problem spanning
task-level mixing, data selection, and curriculum design. During supervised
fine-tuning, mixture-level methods balance transfer against interference and
forgetting across capabilities, while finer-grained methods select or order
instructions according to their relevance, influence, difficulty, or learning
progress
\citep{DBLP:conf/icml/Li0X25,DBLP:conf/emnlp/WuVQH24,DBLP:conf/icml/XiaMGA024,
DBLP:conf/naacl/LiZLCC0W0024,DBLP:conf/icml/Liang00LWCH25}.
During preference optimisation, \citet{DBLP:conf/acl/CorradoKDYZXPYC25}
prioritise tasks on which a generalist model lags behind task-specific
specialists. Related work on reinforcement learning with verifiable rewards
learns data mixtures that improve generalisation across heterogeneous
capabilities \citep{DBLP:journals/corr/abs-2505-24871}.

Across these stages, the utility of training data depends on the target
capability, the base model, and its current training state. We therefore use
\emph{high-quality} in this training-dependent sense: an example is of high
quality for a training run when allocating budget to it improves the target
objective from the current model state. Rule-based filters and external
scorers approximate this utility with fixed, model-independent proxies,
whereas model-aware methods estimate it from losses, gradients, or curvature.
Section~\ref{sec:formulation} makes this utility the objective of a data
policy.


Data strategy is studied under separate names, including selection, pruning,
reweighting, mixture optimisation, and curricula, which obscures their shared
structure and complicates compute comparisons. Existing surveys cover
different parts of this landscape.
\citet{wang2023data} study data management broadly across pretraining and
supervised fine-tuning, while \citet{albalak2024survey} organise language-model
data selection around utility functions and selection mechanisms, with an
emphasis on pretraining. Other surveys focus specifically on instruction
tuning, categorising methods by selector type
\citep{zhang2025survey} or by quality, diversity, and importance
\citep{qin2024tsunami}; Appendix Table~\ref{tab:survey-comparison} compares
their stage coverage. Our contributions are:
\begin{itemize}[leftmargin=*,nosep,topsep=2pt]
    \item We cast selection, reweighting, mixing, and curricula as policy
    optimisation over the training trajectory.
    \item We review methods from pretraining to post-training, comparing
    their assumptions, computational costs, and empirical evidence.
    \item We compare partitions induced by three signals and reconstruct
    auxiliary-training compute for representative data-mixture strategies.
\end{itemize}

\section{Data Strategy Formulation}
\label{sec:formulation}

We formulate data strategies as policies that allocate a training budget over
available examples, then distinguish how selection, reweighting, curricula,
and optimisers constrain those policies.

\subsection{Data Strategies as Policy Optimisation}
\label{sec:formulation:policy}

\noindent
Let $\dataset_t=\{z_{t,i}\}_{i=1}^{n_t}$ be the candidate pool available at
step $t$. In offline settings, $\dataset_t\equiv\dataset$; in online settings,
it may evolve with the model, as with newly generated rollouts. Let
$s_t=(\theta_t,\xi_t,t)\in\mathcal{S}$ be the training state, comprising model
parameters $\theta_t\in\Theta\subseteq\mathbb{R}^{d}$, auxiliary variables
$\xi_t$, and the step index. A data strategy chooses a distribution over the
current pool. With per-example loss $\ell$ and token-count function
$\mathrm{tok}$, we express the resulting optimisation problem as
\begingroup
\setlength{\fboxsep}{2pt}
\begin{equation}
\begin{aligned}
  \min_{\phi\in\Phi}\quad
  &\mathbb{E}\!\bigl[
    \underbrace{\colorbox{objcol!10}{%
      $\displaystyle \objJ(\theta_T)$}}_{
      \smash[b]{\textcolor{objcol}{\text{\scriptsize target objective}}}}
    \bigr]\\[4pt]
  \mathrm{s.t.}\quad
  &\mathbf{q}_t=
    \underbrace{\colorbox{polcol!10}{%
      $\displaystyle \policy(s_t,\dataset_t)$}}_{
      \smash[b]{\textcolor{polcol}{\text{\scriptsize data policy}}}}
    \in\mathcal{Q}_t,\qquad
  s_{t+1}=
    \underbrace{\colorbox{updcol!10}{%
      $\displaystyle \update(s_t,\widehat{\mathbf{g}}_t)$}}_{
      \smash[b]{\textcolor{updcol}{\text{\scriptsize state update}}}},\;
  \textstyle\sum_{t}\mathrm{tok}(\mathcal{B}_t)\leq B.
\end{aligned}
\label{eq:data-policy}
\end{equation}
\endgroup
for $t=0,\dots,T-1$ from a given initial state $s_0$, where $\Phi$ is the
admissible policy class and $\mathcal{Q}_t\subseteq\Delta^{n_t-1}$ encodes its
pointwise constraints. At each step, a batch $\mathcal{B}_t\sim\mathbf{q}_t$ is
sampled from the current pool, and $\widehat{\mathbf{g}}_t=|\mathcal{B}_t|^{-1}
\sum_{z\in\mathcal{B}_t}\nabla_\theta\ell(z;\theta_t)$ is its mean gradient;
$B$ denotes the total token budget.
The expectation is over batch sampling and any online generation of
$\dataset_t$.
Problem~\eqref{eq:data-policy} involves three functions:
\begin{itemize}[leftmargin=*,nosep,topsep=2pt]
\item \textcolor{objcol}{\textbf{Objective function}} $\objJ:\Theta\to\mathbb{R}$
  evaluates the final model, typically by its held-out loss
  $\objJ(\theta)=\mathbb{E}_{z\sim\mathcal{D}_{\mathrm{tgt}}}[\ell(z;\theta)]$
  on a target distribution $\mathcal{D}_{\mathrm{tgt}}$; robust variants are
  discussed in Section~\ref{sec:posttraining:comparison}.
\item \textcolor{polcol}{\textbf{Data policy function}}
  $\policy$ maps the current state and candidate pool to a feasible allocation
  $\mathbf{q}_t$. Zero probabilities encode selection, tied probabilities
  encode group-level mixing, and temporal changes encode scheduling. Here
  $\phi$ collects learnable parameters or design choices.
\item \textcolor{updcol}{\textbf{State-update function}}
  $\update:\mathcal{S}\times\mathbb{R}^{d}\to\mathcal{S}$ takes one
  optimiser step along a stochastic gradient. For Adam, $\xi_t$ contains the
  moment estimates; it may also carry history used by a stateful data policy.
\end{itemize}
The token constraint makes the data budget explicit. Deterministic loss
reweighting is recovered by replacing $\widehat{\mathbf{g}}_t$ with its
conditional expectation
$\sum_i q_{t,i}\nabla_\theta\ell(z_{t,i};\theta_t)$.
For a fixed pool, a state-independent policy yields
$\mathbf{q}_t\equiv\mathbf{q}$. In the idealised convergence limit, this
recovers the familiar bilevel problem
$\min_{\mathbf{q}\in\mathcal{Q}}\objJ(\theta^\star(\mathbf{q}))$
subject to
$\theta^\star(\mathbf{q})\in
\arg\min_{\theta}\sum_i q_i\,\ell(z_i;\theta)$.

\subsection{Disambiguating Related Concepts}
\label{sec:formulation:concepts}

\begin{figure}[t]
\centering
\resizebox{\textwidth}{!}{%
\begin{tikzpicture}[
  x=1cm,
  y=1cm,
  font=\sffamily,
  card/.style={
    rounded corners=2.5mm,
    draw=black!18,
    fill=white,
    line width=0.65pt
  },
  paneltitle/.style={font=\sffamily\small\bfseries},
  panelsubtitle/.style={font=\sffamily\scriptsize, text=black!58},
  formula/.style={font=\sffamily\small, text=black!82},
  sampleoff/.style={
    rounded corners=1.2mm,
    draw=black!22,
    fill=black!2,
    minimum width=0.72cm,
    minimum height=0.55cm,
    inner sep=0pt,
    font=\sffamily\scriptsize,
    text=black!48
  },
  sampleon/.style={
    rounded corners=1.2mm,
    draw=gencol!80,
    fill=gencol!9,
    line width=0.75pt,
    minimum width=0.72cm,
    minimum height=0.55cm,
    inner sep=0pt,
    font=\sffamily\scriptsize\bfseries,
    text=gencol
  }
]

\node[
  fill=white,
  inner xsep=7pt,
  font=\sffamily\footnotesize\bfseries,
  text=black!65
] at (8.25,5.10) {distinct decision units and allocation constraints};

\draw[card] (0,0) rectangle (5.20,4.70);
\draw[findingblue, line width=2.2pt, line cap=round]
  (0.28,4.66) -- (4.92,4.66);
\node[circle, fill=findingblue, text=white, inner sep=2.3pt,
      font=\sffamily\scriptsize\bfseries] at (0.45,4.25) {1};
\node[paneltitle, text=findingblue, anchor=west]
  at (0.82,4.25) {Dataset-level selection};
\node[panelsubtitle] at (2.60,3.82) {unit: an entire dataset};
\node[
  rounded corners=4pt,
  draw=findingblue!28,
  fill=findingblue!5,
  inner xsep=6pt,
  inner ysep=1.5pt,
  font=\sffamily\scriptsize,
  text=findingblue
] at (2.60,3.42) {budget $k=2$};

\node[font=\sffamily\scriptsize\bfseries, text=findingblue]
  at (1.37,3.06) {$\mathcal{C}_1$};
\draw[rounded corners=1.2mm, draw=sftcol!85, fill=sftcol!7,
      line width=0.75pt] (0.60,2.02) rectangle (2.14,2.91);
\foreach \yy in {2.25,2.47,2.69} {
  \fill[sftcol!65] (0.82,\yy) circle (0.055);
  \draw[black!43, line width=0.45pt] (0.98,\yy) -- (1.88,\yy);
}
\node[circle, fill=sftcol, text=white, inner sep=1.2pt,
      font=\sffamily\tiny\bfseries] at (2.04,2.82) {\ding{51}};

\node[font=\sffamily\scriptsize\bfseries, text=black!48]
  at (3.83,3.06) {$\mathcal{C}_2$};
\draw[rounded corners=1.2mm, draw=black!22, fill=black!2]
  (3.06,2.02) rectangle (4.60,2.91);
\foreach \yy in {2.25,2.47,2.69} {
  \fill[black!28] (3.28,\yy) circle (0.055);
  \draw[black!28, line width=0.45pt] (3.44,\yy) -- (4.34,\yy);
}
\node[circle, fill=objcol, text=white, inner sep=1.2pt,
      font=\sffamily\tiny\bfseries] at (4.50,2.82) {\ding{55}};

\node[font=\sffamily\scriptsize\bfseries, text=black!48]
  at (1.37,1.90) {$\mathcal{C}_3$};
\draw[rounded corners=1.2mm, draw=black!22, fill=black!2]
  (0.60,0.86) rectangle (2.14,1.75);
\foreach \yy in {1.09,1.31,1.53} {
  \fill[black!28] (0.82,\yy) circle (0.055);
  \draw[black!28, line width=0.45pt] (0.98,\yy) -- (1.88,\yy);
}
\node[circle, fill=objcol, text=white, inner sep=1.2pt,
      font=\sffamily\tiny\bfseries] at (2.04,1.66) {\ding{55}};

\node[font=\sffamily\scriptsize\bfseries, text=findingblue]
  at (3.83,1.90) {$\mathcal{C}_4$};
\draw[rounded corners=1.2mm, draw=sftcol!85, fill=sftcol!7,
      line width=0.75pt] (3.06,0.86) rectangle (4.60,1.75);
\foreach \yy in {1.09,1.31,1.53} {
  \fill[sftcol!65] (3.28,\yy) circle (0.055);
  \draw[black!43, line width=0.45pt] (3.44,\yy) -- (4.34,\yy);
}
\node[circle, fill=sftcol, text=white, inner sep=1.2pt,
      font=\sffamily\tiny\bfseries] at (4.50,1.66) {\ding{51}};
\node[formula] at (2.60,0.39)
  {$\mathbf{b}^{\mathrm{set}}\in\{0,1\}^{M},\quad
    \lVert\mathbf{b}^{\mathrm{set}}\rVert_0\leq k$};

\draw[card] (5.65,0) rectangle (10.85,4.70);
\draw[updcol, line width=2.2pt, line cap=round]
  (5.93,4.66) -- (10.57,4.66);
\node[circle, fill=updcol, text=white, inner sep=2.3pt,
      font=\sffamily\scriptsize\bfseries] at (6.10,4.25) {2};
\node[paneltitle, text=updcol, anchor=west]
  at (6.47,4.25) {Domain reweighting};
\node[panelsubtitle] at (8.25,3.82) {unit: a predefined domain};

\node[font=\sffamily\scriptsize, anchor=east, text=black!68]
  at (6.95,3.25) {Code};
\draw[rounded corners=1pt, fill=black!6, draw=none]
  (7.20,3.13) rectangle (9.70,3.37);
\draw[rounded corners=1pt, fill=updcol, draw=none]
  (7.20,3.13) rectangle (9.70,3.37);
\node[font=\sffamily\scriptsize\bfseries, anchor=west, text=updcol]
  at (9.88,3.25) {$0.35$};

\node[font=\sffamily\scriptsize, anchor=east, text=black!68]
  at (6.95,2.63) {Math};
\draw[rounded corners=1pt, fill=black!6, draw=none]
  (7.20,2.51) rectangle (9.70,2.75);
\draw[rounded corners=1pt, fill=updcol!82, draw=none]
  (7.20,2.51) rectangle (9.34,2.75);
\node[font=\sffamily\scriptsize\bfseries, anchor=west, text=updcol]
  at (9.88,2.63) {$0.30$};

\node[font=\sffamily\scriptsize, anchor=east, text=black!68]
  at (6.95,2.01) {Web};
\draw[rounded corners=1pt, fill=black!6, draw=none]
  (7.20,1.89) rectangle (9.70,2.13);
\draw[rounded corners=1pt, fill=updcol!62, draw=none]
  (7.20,1.89) rectangle (8.63,2.13);
\node[font=\sffamily\scriptsize\bfseries, anchor=west, text=updcol]
  at (9.88,2.01) {$0.20$};

\node[font=\sffamily\scriptsize, anchor=east, text=black!68]
  at (6.95,1.39) {Books};
\draw[rounded corners=1pt, fill=black!6, draw=none]
  (7.20,1.27) rectangle (9.70,1.51);
\draw[rounded corners=1pt, fill=updcol!42, draw=none]
  (7.20,1.27) rectangle (8.27,1.51);
\node[font=\sffamily\scriptsize\bfseries, anchor=west, text=updcol]
  at (9.88,1.39) {$0.15$};

\node[font=\sffamily\scriptsize, text=black!48]
  at (8.25,0.86) {sampling proportions};
\node[formula] at (8.25,0.39)
  {$\boldsymbol{\alpha}\in\Delta^{K-1},\quad \sum_k\alpha_k=1$};

\draw[card] (11.30,0) rectangle (16.50,4.70);
\draw[gencol, line width=2.2pt, line cap=round]
  (11.58,4.66) -- (16.22,4.66);
\node[circle, fill=gencol, text=white, inner sep=2.3pt,
      font=\sffamily\scriptsize\bfseries] at (11.75,4.25) {3};
\node[paneltitle, text=gencol, anchor=west]
  at (12.12,4.25) {Example-level selection};
\node[panelsubtitle] at (13.90,3.82) {unit: one training example};
\node[
  rounded corners=4pt,
  draw=gencol!28,
  fill=gencol!5,
  inner xsep=6pt,
  inner ysep=1.5pt,
  font=\sffamily\scriptsize,
  text=gencol
] at (13.90,3.42) {6 of 15 retained};

\node[sampleon]  at (11.85,2.78) {$z_1$};
\node[sampleoff] at (12.85,2.78) {$z_2$};
\node[sampleon]  at (13.85,2.78) {$z_3$};
\node[sampleoff] at (14.85,2.78) {$z_4$};
\node[sampleon]  at (15.85,2.78) {$z_5$};

\node[sampleoff] at (11.85,2.05) {$z_6$};
\node[sampleon]  at (12.85,2.05) {$z_7$};
\node[sampleoff] at (13.85,2.05) {$z_8$};
\node[sampleon]  at (14.85,2.05) {$z_9$};
\node[sampleoff] at (15.85,2.05) {$z_{10}$};

\node[sampleoff] at (11.85,1.32) {$z_{11}$};
\node[sampleoff] at (12.85,1.32) {$z_{12}$};
\node[sampleoff] at (13.85,1.32) {$z_{13}$};
\node[sampleon]  at (14.85,1.32) {$z_{14}$};
\node[sampleoff] at (15.85,1.32) {$z_{15}$};

\node[font=\sffamily\scriptsize, text=black!48]
  at (13.90,0.86) {binary decisions per example};
\node[formula] at (13.90,0.39)
  {$\mathbf{b}^{\mathrm{ex}}\in\{0,1\}^{n}$};

\end{tikzpicture}%
}
\caption{\textbf{Decision granularity in LLM data strategies.}
Dataset-level selection retains whole corpora, domain reweighting assigns
continuous weights to predefined groups, and example-level selection retains
individual examples. Their feasible sets need not be nested.}
\label{fig:selection-granularity}
\end{figure}
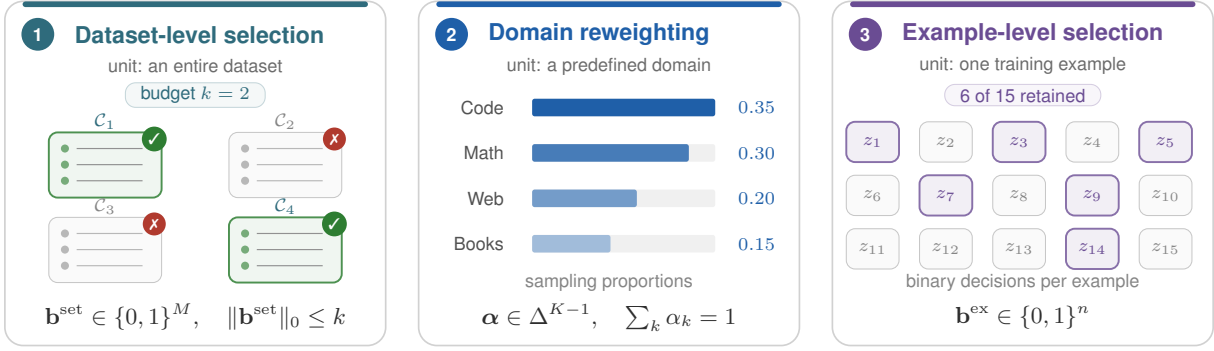

\noindent
The formulation separates the roles of optimisers, selection, reweighting,
and curricula. We relate their decision spaces and compare their use across
training stages. Figure~\ref{fig:selection-granularity} visualises the
corresponding granularities.

\keypoint{ptcol}{How do optimisers, selection, reweighting, and curricula
differ?}

\paragraph{Optimisers and data policies.}
Both modify gradient contributions, but along different axes. A data policy
assigns scalar probabilities to examples available at the current state,
whereas Adam applies coordinate-wise coefficients to historical batch
gradients. They differ in the scope of their decisions: a data policy changes
the current training distribution, whereas Adam transforms accumulated
gradients inside $\update$. More explicitly, a data policy assigns scalar
probabilities to current examples, whereas Adam applies coordinate-wise
coefficients to historical batch gradients:
\begingroup
\setlength{\fboxsep}{2pt}
\begin{equation*}
\left\{
\begin{aligned}
\mathbf{g}_t^{\mathrm{data}}
  &= \sum_{j=1}^{n_t}
    \underbrace{\colorbox{ptcol!10}{%
      $\displaystyle [\pi_\phi(s_t,\dataset_t)]_j$}}_{
      \smash[b]{\textcolor{ptcol}{q^{\mathrm{data}}_{t,j}}}}
    \mathbf{g}_{t,j},\\[8pt]
\mathbf{g}_t^{\mathrm{Adam}}
  &= \sum_{j=1}^{t}
    \underbrace{\colorbox{prefcol!10}{%
      $\displaystyle
      \frac{a_{t,j}}{
        \sqrt{\sum_{k=1}^{t}b_{t,k}\mathbf{h}_k^{\odot 2}}+\epsilon}$}}_{
      \smash[b]{\textcolor{prefcol}{\mathbf{w}^{\mathrm{Adam}}_{t,j}}}}
    \odot\mathbf{h}_j .
\end{aligned}
\right.
\end{equation*}
\endgroup
Here $\mathbf{g}_{t,j}=\nabla_\theta\ell(z_{t,j};\theta_t)$ is a current
example gradient, while
$\mathbf{h}_j=\nabla_\theta\ell(\mathcal{B}_j;\theta_{j-1})$ is the gradient
of a previously observed batch. Moreover,
$a_{t,j}=(1-\beta_1)\beta_1^{t-j}/(1-\beta_1^t)$ and
$b_{t,k}=(1-\beta_2)\beta_2^{t-k}/(1-\beta_2^t)$.

\paragraph{Selection versus reweighting.}
Data reweighting permits any allocation in the simplex. Data selection is the
special case that assigns equal probability to retained examples and zero to
the rest; a fixed cardinality budget further imposes $|S|=k$. Domain
reweighting constrains all examples in a domain to share one per-example
probability. This reduces the decision from $n$ probabilities to $K$ mixture
proportions, making it tractable at pretraining scale while preventing the
policy from isolating one example within a multi-example domain.
Section~\ref{sec:pretraining:reweighting} reviews how prior work defines
domains.

\begin{proposition}[Decision-space constraints]
\label{prop:hierarchy}
Assume $n\geq 2$. Partition $\dataset$ into non-empty domains
$\dataset_1,\dots,\dataset_K$ and let
$A\in\mathbb{R}^{n\times K}$ with $A_{ik}=1/|\dataset_k|$ if
$z_i\in\dataset_k$ and $A_{ik}=0$ otherwise. The feasible sets of
data reweighting, data selection, and domain reweighting are
$\mathcal{Q}_{\mathrm{rw}}=\Delta^{n-1}$,
$\mathcal{Q}_{\mathrm{ds}}=\{\mathbf{1}_S/|S|:\emptyset\neq S\subseteq[n]\}$,
and $\mathcal{Q}_{\mathrm{dr}}=\{A\alpha:\alpha\in\Delta^{K-1}\}$. Then
$\mathcal{Q}_{\mathrm{ds}}\subsetneq\mathcal{Q}_{\mathrm{rw}}$ and
$\mathcal{Q}_{\mathrm{dr}}\subseteq\mathcal{Q}_{\mathrm{rw}}$, with equality
in the latter if and only if $K=n$. If $K\geq 2$ and
some domain contains at least two examples, neither
$\mathcal{Q}_{\mathrm{ds}}$ nor $\mathcal{Q}_{\mathrm{dr}}$ contains the other.
\end{proposition}
\noindent The inclusions follow directly from the simplex definitions, and
$\mathcal{Q}_{\mathrm{dr}}=\mathcal{Q}_{\mathrm{rw}}$ only when every domain
is a singleton. For incomparability, selecting one member of a multi-example
domain violates the domain constraint, while unequal positive per-example
probabilities across two domains violate the uniform-support constraint of
selection. Data selection is therefore not, in general, a discretisation of
domain reweighting.

\paragraph{Curricula versus selection and reweighting.}
A curriculum is a structured data policy whose allocation factors through an
explicit curriculum signal and a scheduling rule. It constrains the trajectory
$(\mathbf{q}_0,\dots,\mathbf{q}_{T-1})$ rather than only the feasible set at
each step, and can be combined with selection, example reweighting, or domain
reweighting. A static curriculum fixes its signal and schedule before
training; an adaptive curriculum updates either component from the current
state.
Section~\ref{sec:curriculum-policy} formalises this policy class and compares
the signals and scheduling rules used in prior work.
\keypoint{ptcol}{How do data strategies differ between pretraining and
post-training?}
\noindent Both stages instantiate Equation~\eqref{eq:data-policy}, while
differing in scale, supervision, pool dynamics, and model state. Standard
pretraining optimises next-token prediction over massive, largely static
corpora, favouring coarse domain-level decisions or infrequent rescoring.
Annealing and mid-training begin from an existing checkpoint but retain many
of the same scale constraints. Post-training uses instruction, preference, or
reward feedback. Its pool may be fixed for supervised fine-tuning and offline
preference optimisation, or evolve through model-generated rollouts in online
reinforcement learning. Smaller pools and richer feedback often make
fine-grained, model-conditioned policies more practical.

\subsection{Curricula as Structured Policies}
\label{sec:curriculum-policy}

We represent a curriculum by a scalar signal
$r_\psi(z_{t,i},s_t)$ and a scheduling rule
$h_\omega(r,c_t)\geq 0$, where $c_t$ is a progression variable. The signal and
rule belong to prespecified hypothesis classes $\Psi$ and $\Omega$. The
resulting allocation is
\begin{equation}
\begin{aligned}
\widetilde q_{t,i}
  &= h_\omega\!\left(r_\psi(z_{t,i},s_t),c_t\right),\\
q_{t,i}
  &= \widetilde q_{t,i}
     \Big/ \sum_{j=1}^{n_t}\widetilde q_{t,j},
\qquad i=1,\ldots,n_t,
\end{aligned}
\label{eq:curriculum}
\end{equation}
where $\psi\in\Psi$, $\omega\in\Omega$, and
$\sum_j\widetilde q_{t,j}>0$. If $\Psi$ and $\Omega$ exclude example- and
time-indexed lookup tables, this factorisation defines a structured subclass
of data policies. In the offline setting, a static curriculum fixes
$r_\psi(z_i,s_0)$ and $(c_0,\ldots,c_{T-1})$ before training. An adaptive
curriculum instead updates the signal or progression variable from the
current state, for example $c_t=g(s_t)$.

\paragraph{Criteria and scheduling.}
Equation~\eqref{eq:curriculum} separates the scheduled unit, the signal used
to value it, and the rule that changes its exposure. Recent methods schedule
examples, clusters, data categories, prompts, or replayed prompt--output pairs
\citep{DBLP:journals/corr/abs-2405-07490,DBLP:conf/icml/Liang00LWCH25,
pang2026edcodynamiccurriculumorchestration,do-etal-2026-space,
kim2026normalizationrethinkingpartitionfunction}. SPARD additionally adapts
reward-objective weights, a decision outside the allocation
$\mathbf q_t$~\citep{zhi-etal-2026-spard}.

\paragraph{Static and adaptive curricula.}
Strategic Ordering uses a random first epoch, then fixes an easy-to-hard order
from prompt length, attention, or loss for later
epochs~\citep{DBLP:journals/corr/abs-2405-07490}. EDCO periodically ranks
examples with the authors' prefix negative-log-likelihood score, termed
inference entropy, and selects the top-$N$ examples for the next
phase~\citep{pang2026edcodynamiccurriculumorchestration}. It combines data
selection with an adaptive curriculum under Equation~\eqref{eq:curriculum}.

\paragraph{Beyond easy-to-hard ordering.}
Recent methods increasingly replace fixed difficulty proxies with
state-dependent estimates of learning value, including gradient interaction,
inference entropy, solve rate, reward progress, and online
accuracy~\citep{DBLP:conf/icml/Liang00LWCH25,
pang2026edcodynamiccurriculumorchestration,do-etal-2026-space,
zhi-etal-2026-spard,kim2026normalizationrethinkingpartitionfunction}.
Their schedules need not be easy-to-hard. EDCO prioritises high uncertainty,
PACED-RL targets intermediate difficulty, and SPARD favours dimensions with
the largest reliable recent gains~\citep{pang2026edcodynamiccurriculumorchestration,
kim2026normalizationrethinkingpartitionfunction,zhi-etal-2026-spard}. Thus,
``curriculum'' identifies temporal control of exposure, not a particular
notion of difficulty or a mandatory easy-to-hard direction.

\section{Data Strategies for Pretraining}
\label{sec:pretraining}

\subsection{A Taxonomy of Methods}
\label{sec:pretraining:taxonomy}

Pretraining data strategies differ mainly in the granularity they act on: data
selection decides \emph{which documents} are retained for training, whereas
domain reweighting decides \emph{how much} of each domain to sample.

\subsubsection{Data Selection}
\label{sec:pretraining:selection}

\keypoint{ptcol}{How is the value of an individual document judged?}
\noindent Data selection judges each individual document and decides whether
to retain it for training. By the source of
this judgement, existing methods fall into two families, rule-based filtering
and model-based selection. The latter further divides into scoring with an
external model and signals read from the target model itself.

\begin{table}[H]
\centering
\footnotesize
\setlength{\tabcolsep}{3.5pt}
\renewcommand{\arraystretch}{1.2}
\newcommand{\rulegroup}[2]{\rowcolor{rootcol!10}\multicolumn{3}{@{}l@{}}{\textcolor{rootcol}{\textbf{#2}}}\\}
\newcommand{\src}[1]{\textcolor{rootcol}{#1}}
\newcommand{\ricon}[2]{\makebox[1.5em][c]{\textcolor{#1}{\scriptsize#2}}\,}
\tablestripes
\begin{tabular}{@{}L{0.31\columnwidth}L{0.42\columnwidth}L{0.20\columnwidth}@{}}
\toprule
\tableheadrow
\hspace{1.67em}\textbf{Rule type} & \textbf{Example rule} & \textbf{Source}\\
\midrule
\rulegroup{gencol}{Metadata only}
\ricon{gencol}{\faLink}URL / domain & Domain not on a blocklist          & \src{RefinedWeb}\\
\rulegroup{ptcol}{Single document: statistic + threshold}
\ricon{ptcol}{\faRulerHorizontal}Size & $50$--$100$K words                 & \src{Gopher}\\
\ricon{ptcol}{\faFont}Composition & Mean word length $3$--$10$         & \src{Gopher}\\
\ricon{ptcol}{\faAlignLeft}Line structure & Punctuated lines ${>}12\%$ & \src{FineWeb}\\
\ricon{ptcol}{\faRedo}Repetition & ${\le}30\%$ duplicated lines       & \src{Gopher}\\
\rulegroup{prefcol}{Single document: pattern matching}
\ricon{prefcol}{\faSpellCheck}Lexicon & No listed bad words      & \src{C4}\\
\rulegroup{sftcol}{Across documents}
\ricon{sftcol}{\faClone}Deduplication & No near-duplicates          & \src{FineWeb}\\
\ricon{sftcol}{\faChartBar}Corpus prior & Mean token prior in range   & \src{Prior-based}\\
\bottomrule
\end{tabular}
\caption{\textbf{Rule-based filters.} Filters are grouped by the information
each rule requires. Thresholds follow the cited source. FineWeb2 re-derives
them for each language.}
\label{tab:rule-filters}
\end{table}

\paragraph{Rule-based filtering.}
Rule-based methods apply hand-crafted heuristics that require no learned model.
Table~\ref{tab:rule-filters} groups them by the information a rule needs:
metadata such as the URL, statistics or patterns within a single document, or
comparisons across the corpus. Most open corpora, such as
RefinedWeb~\citep{penedo2023refinedweb} and Dolma~\citep{soldaini2024dolma},
combine the rules of C4~\citep{raffel2020c4} and Gopher~\citep{rae2021gopher}
with exact or MinHash deduplication.
FineWeb ablates these rules one by one and replaces C4's terminal-punctuation
rule, which removes about $30\%$ of tokens, with milder custom
filters~\citep{penedo2024fineweb}. Recent work makes the rules more adaptive:
FineWeb2 derives thresholds for each language from that language's metric
distributions~\citep{penedo2025fineweb2}, and prior-based filtering replaces
model perplexity with corpus-level token frequencies, cutting filtering time by
more than $1000\times$~\citep{seo2026priorfilter}. Such rules are cheap and
transparent, but they capture only surface properties of a document rather than
its value for training, and their thresholds must be re-tuned for each corpus
or language.

\paragraph{External-model scoring.}
Model-based selection replaces hand-crafted rules with a learned judgement.
One branch uses a separate model to score each document and retains those with
high scores (Table~\ref{tab:external-scorers}). Quality can be defined by a
reference signal: DataComp-LM (DCLM) trains a classifier to separate curated
data from generic web text~\citep{li2024dclm}, and PreSelect favours documents
whose losses across a set of models best predict those models' downstream
ability~\citep{shum2025preselect}. It can
also be defined by an LLM, either through rated properties, such as
educational value in FineWeb-Edu~\citep{penedo2024fineweb}, writing style and
expertise in QuRating~\citep{wettig2024qurating}, and professionalism,
readability, reasoning, and cleanliness (PRRC) in
Meta-rater~\citep{DBLP:conf/acl/ZhuangPMWBWQZQH25}, or through a direct
judgement of whether a document is useful for pretraining, as in
AskLLM~\citep{sachdeva2024askllm}. Classifiers also flag content that is
undesirable regardless of quality, such as toxic text in
Dolma~\citep{soldaini2024dolma} or sensitive personal
information~\citep{minamoto-etal-2026-detecting}. QuaDMix combines
several such scores with domain labels to balance quality and
diversity~\citep{DBLP:journals/corr/abs-2504-16511}.

The main limitation is
that these scorers are independent of the model being trained, so the quality
they encode may not match what the target model needs. The rated dimensions
also differ in value. In QuRating, selecting by writing style gives the lowest
perplexity but little downstream gain over uniform
selection~\citep{wettig2024qurating}, and Meta-rater assigns its largest
learned weight to educational value~\citep{DBLP:conf/acl/ZhuangPMWBWQZQH25}.
Scorers are also sensitive to surface style. Rephrasing documents in Wikipedia
style reverses the FineWeb-Edu classifier's decision for about $7\%$ of
them~\citep{klimaszewski2026cqfpitfalls}. Classifier-based filtering can
improve downstream accuracy without lowering the loss on the high-quality set
it is trained to imitate, because the selected data resemble a filtered subset
of that set, not the set as a
whole~\citep{saada2026removingnoisefindinggold}. Each new dimension also
requires fresh annotations, usually from an LLM, before a scorer can be
trained.

\begin{table}[t]
\centering
\footnotesize
\setlength{\tabcolsep}{3.5pt}
\renewcommand{\arraystretch}{1.2}
\newcommand{\rulegroup}[2]{\rowcolor{rootcol!10}\multicolumn{3}{@{}l@{}}{\textcolor{rootcol}{\textbf{#2}}}\\}
\newcommand{\src}[1]{\textcolor{rootcol}{#1}}
\newcommand{\ricon}[2]{\makebox[1.5em][c]{\textcolor{#1}{\scriptsize#2}}\,}
\tablestripes
\begin{tabular}{@{}L{0.33\columnwidth}L{0.37\columnwidth}L{0.23\columnwidth}@{}}
\toprule
\tableheadrow
\hspace{1.67em}\textbf{Dimension} & \textbf{Scorer} & \textbf{Source}\\
\midrule
\rulegroup{gencol}{Quality from a reference signal}
\ricon{gencol}{\faBullseye}Similarity & Curated vs.\ web & \src{DCLM}\\
\ricon{gencol}{\faChartLine}Predictiveness & Loss--benchmark fit & \src{PreSelect}\\
\rulegroup{ptcol}{Quality defined by an LLM}
\ricon{ptcol}{\faGraduationCap}Education & Distilled LLM labels & \src{FineWeb-Edu}\\
\ricon{ptcol}{\faPenNib}Style / expertise & Pairwise LLM ratings & \src{QuRating}\\
\ricon{ptcol}{\faBrain}PRRC & Learned combination & \src{Meta-rater}\\
\ricon{ptcol}{\faComments}Usefulness & Prompted LLM & \src{AskLLM}\\
\rulegroup{sftcol}{Undesirable content}
\ricon{sftcol}{\faBan}Toxicity & Bigram classifier & \src{Dolma}\\
\ricon{sftcol}{\faUserShield}Privacy & SCPI classifier & \src{Minamoto et al.}\\
\bottomrule
\end{tabular}
\caption{\textbf{External-model scorers.} Scorers are grouped by how the
target property is defined. PRRC denotes professionalism, readability, reasoning, and
cleanliness; SCPI denotes sensitive personal information.}
\label{tab:external-scorers}
\end{table}

\paragraph{Target-model signals.}
The other branch reads selection signals from the target model, a smaller
proxy of it, or an off-the-shelf LLM. We group these methods by the order of information they
extract from the training loss.

Zeroth-order methods score data from forward-pass quantities, such as losses
or activations, without per-sample gradients.
AttentionInfluence ranks documents by the loss increase caused by masking
retrieval attention heads~\citep{DBLP:journals/corr/abs-2505-07293}. This
mechanistic score needs neither a trained scorer nor a loss gradient and
transfers from a $1.3$B selector to a $7$B target.
Ranking based on neuron-activated graphs (NAG) instead compares the sparse sets of high-impact neurons that
candidate and target examples
activate~\citep{wang2026targetorientedpretrainingdataselection}. At a finer,
span-level granularity, PretrainZero regards a span as valuable when it is
informative and not yet mastered by the current model, but remains recoverable
from context; an on-policy mask generator turns such spans into
self-verifiable reinforcement learning
targets~\citep{xing2025pretrainzeroreinforcementactivepretraining}.

First-order methods score a document by its per-sample gradient,
typically through its alignment with a target direction. DsDm models target
performance as a function of the training subset with
datamodels~\citep{DBLP:conf/icml/EngstromFM24}. GREATS greedily selects, at
every iteration, the batch that most reduces a validation loss under a Taylor
approximation~\citep{DBLP:conf/nips/WangWSM024}. OPUS projects each candidate's
optimiser-shaped update, rather than its raw gradient, onto a target
direction~\citep{wang2026opusefficientprincipleddata}. Grad-Mimic measures
how well a sample's gradient aligns with a direction induced by a pretrained
reference model~\citep{huang2026evaluatingsampleutilityefficient}.

Second-order methods further account for curvature through the Hessian.
Pontryagin maximum principle-based data selection (PDS) casts selection as an
optimal-control problem whose costate update involves the Hessian, and distils
the resulting scores into a scorer~\citep{DBLP:conf/iclr/GuDWHDWH25}. BLISS
learns influence scores through bilevel optimisation with Hessian-vector
products~\citep{DBLP:journals/corr/abs-2510-06048}. DiReCT constrains
per-sample gradients along the eigen-directions of the Hessian during
annealing~\citep{xu2026direct}. Dynamic directional graph-constrained data
scheduling (D$^3$) uses a second-order expansion to model asymmetric influence
between training units, then schedules data over the resulting directed
graph~\citep{xu2026d3dynamicdirectionalgraphconstrained}. Higher-order signals
capture interactions that lower-order ones miss, but are more expensive to
compute.

\subsubsection{Domain Reweighting}
\label{sec:pretraining:reweighting}

We organise domain-reweighting methods around two aspects: the domain
definitions adopted in prior work and the approaches used to solve the
resulting mixture-optimisation problem.

\keypoint{ptcol}{How do existing studies define a domain for LLM training?}
\noindent A domain is not an intrinsic attribute of a data point. Operationally,
it is one cell in a partition $\mathcal{P}=\{\dataset_k\}_{k=1}^{K}$ of the
pretraining corpus $\dataset$ and therefore the unit to which a mixture weight
$w_k$ is assigned. Existing work constructs this partition using three types
of signal.

\paragraph{Human-defined domains.}
Some studies partition a pretraining corpus using human-specified attributes,
most commonly data
source~\citep{DBLP:conf/nips/Xie0DDLLLL0Y23,DBLP:conf/icml/FanPJ24,DBLP:conf/iclr/LiuZMZDPJL25,DBLP:journals/corr/abs-2502-06244,kang2024autoscale}
or topic~\citep{peng2025topicsource,DBLP:journals/corr/abs-2504-00695}.
These domains are easy to interpret and inspect, but they have two
limitations. First, a human-defined taxonomy may not align with the
distinctions that matter to the model. Corpora considered similar by humans
may induce different learning dynamics, whereas nominally different domains
may provide redundant training
signals~\citep{peng2025topicsource,xu-etal-2026-rethinking}. Second, some
corpora resist meaningful assignment
to a single domain. Web-derived corpora such as Pile-CC, for example, contain
heterogeneous topics, genres, and quality levels despite being treated as one
corpus-level domain. Topic-modelling methods can reduce reliance on manually
specified labels~\citep{peng2025topicsource,DBLP:journals/corr/abs-2504-00695},
but the resulting partitions still depend on modelling assumptions and
assignment errors.

\paragraph{Semantic-space partitions.}
Other studies regard semantically close data as a domain and typically rely on
encoder models to embed the
corpus~\citep{diao2025nemotronclimb,DBLP:conf/icml/ZhangTWQ25,peng2025topicsource}.
For example,
Domain2Vec clusters document embeddings into distinct meta-domains and
represents each source corpus as their
distribution~\citep{DBLP:conf/icml/ZhangTWQ25}. A single source-defined domain
can therefore span multiple semantic domains. Nevertheless,
Domain2Vec still optimises mixture weights over the original source corpora.

\paragraph{Gradient-space partitions.}
The preceding two definitions are independent of the language model being
trained, since their partitions come from human labels or an external
encoder~\citep{xu-etal-2026-rethinking}. In contrast, gradient-space methods
construct domains from the target
model or an architecturally similar proxy, grouping samples whose gradients
induce similar parameter-update
directions~\citep{DBLP:conf/emnlp/ShiZWFZZCZCZHXZ25,xu-etal-2026-rethinking}.
This construction requires
computing and storing per-sample gradients.
Domain impact-aware data sampling (DIDS) empirically shows that gradient-based
partitions can transfer across model
scales, but provides no rigorous theoretical guarantee that the underlying
gradient signals are scale
invariant~\citep{DBLP:conf/emnlp/ShiZWFZZCZCZHXZ25}.

Existing comparisons do not establish a generally preferable domain
definition. Appendix~\ref{sec:empirical:rq1} compares source-, semantic-, and
gradient-based partitions of the same documents.

The partition and the mixture optimiser are separate design choices. Fixing
$\mathcal P$ before optimising $w$ assumes that examples within each domain
can share one weight. A better optimiser cannot recover distinctions erased
by that partition, yet most studies optimise weights over a fixed taxonomy
rather than revising the partition and mixture jointly.


\keypoint{ptcol}{How is the mixture-optimisation problem solved?}
\noindent Directly optimising Problem~\eqref{eq:data-policy} would require many
full pretraining runs. Methods either learn a surrogate of the map from mixture
weights to the objective function $\objJ$, or reformulate the objective so that
it can be optimised without full training runs.
Table~\ref{tab:mixture-models} compares representative surrogate classes and
solvers.

\begin{table}[t]
\centering
\footnotesize
\setlength{\tabcolsep}{3pt}
\renewcommand{\arraystretch}{1.25}
\tablestripes
\begin{tabular}{@{}L{0.21\columnwidth}L{0.17\columnwidth}L{0.30\columnwidth}L{0.23\columnwidth}@{}}
\toprule
\tableheadrow
\textbf{Method} & \textbf{Class} & \textbf{Surrogate} $\widehat{\objJ}(w)$ & \textbf{Solver}\\
\midrule
RegMix & Linear & $b_0+b^\top w$ & Random search\\
MoDoMoDo & Quadratic & $a+b^\top w+w^\top Cw$ & Grid search\\
Data Mixing Laws & Log-linear & $c_i+k_i\exp(t_i^\top w)$ & Grid search\\
DUET & Gaussian process & $\sim\mathcal{GP}(\mu,\kappa)$ & Bayesian opt.\\
\midrule
Data Mixing Agent & Neural policy & $\mathbf{w}_t=\policy(s_t)$ & Offline RL\\
\bottomrule
\end{tabular}
\caption{\textbf{Performance modelling for domain reweighting.} Surrogate
class and solver are separate design choices. The first four methods fit
$\widehat{\objJ}$ over static mixtures, whereas Data Mixing Agent learns a
state-dependent policy. MoDoMoDo is included for its surrogate class.}
\label{tab:mixture-models}
\end{table}

\paragraph{Performance modelling.}
These methods restrict the data policy function $\policy$ to an offline
mixture $w\in\Delta^{K-1}$ and treat the rest of
Problem~\eqref{eq:data-policy}, namely running the state-update function $\update$
for $T$ steps and scoring the result with $\objJ$, as a black box
$w\mapsto\objJ(\theta_T(w))$. Because evaluating this map at full scale for
many mixtures is prohibitively expensive, they evaluate a set of candidate
mixtures, usually with small proxy models, and fit a surrogate
$\widehat{\objJ}(w)$ to the observed objective values. The model class fixes the form of
$\widehat{\objJ}$. RegMix uses linear models or gradient-boosted
trees~\citep{DBLP:conf/iclr/LiuZMZDPJL25}, MoDoMoDo uses a quadratic
form~\citep{DBLP:journals/corr/abs-2505-24871}, Data Mixing Laws uses a
log-linear form~\citep{DBLP:conf/iclr/YeLSZZQ25}, and DUET uses a Gaussian
process~\citep{DBLP:journals/corr/abs-2502-00270}. A solver, such as random
search, grid search, or Bayesian optimisation, then finds
$w^\star\in\arg\min_{w\in\Delta^{K-1}}\widehat{\objJ}(w)$, and the selected
mixture is used for full-scale training. Data Mixing Agent departs from this
pattern by learning a state-dependent $\policy$ directly with offline
reinforcement learning~\citep{DBLP:journals/corr/abs-2507-15640}.

The fitted optimum must transfer twice: from observed to unobserved mixtures,
and from proxy runs to the target scale. RegMix reports stable mixture rankings
across the scales it tests~\citep{DBLP:conf/iclr/LiuZMZDPJL25}, whereas
AutoScale models how the optimum changes with token
budget~\citep{kang2024autoscale}. Neither approach guarantees transfer beyond
its sampled model and data regimes.

\paragraph{Optimisation reformulation.}
A second class of methods replaces the objective with one that is cheaper to
evaluate or optimise. DoReMi replaces the terminal objective $\objJ(\theta_T)$
with the worst domain-wise excess loss of a small proxy model relative to a
reference model. Because this surrogate can be evaluated at every step,
$\policy$ becomes an online policy whose domain weights are updated alternately
with the proxy parameters; DoReMi averages these weights over training and
uses them as a static mixture for the full-scale
model~\citep{DBLP:conf/nips/Xie0DDLLLL0Y23}. MixMin instead replaces
Problem~\eqref{eq:data-policy} with a convex objective over domain mixtures that
can be solved directly~\citep{DBLP:conf/icml/ThudiRRTM25}. Aioli makes the
objective local in time. Within each round of the target run, it estimates a
linear mixing law from short interleaved training intervals and moves the
mixture to reduce the next-round validation
loss~\citep{DBLP:conf/iclr/ChenHLCR25}.

\subsection{Summary and Discussion}
\label{sec:pretraining:summary}

\paragraph{Trend.}
More data selection methods now estimate the value of data relative to a
model and a target objective, using losses, gradients, or
curvature~\citep{DBLP:conf/icml/EngstromFM24,DBLP:conf/nips/WangWSM024,
DBLP:conf/iclr/GuDWHDWH25}. Rules and external scorers, which treat quality
as a property of the text itself, are still the standard way to build large
open corpora because they are cheaper to apply at corpus
scale~\citep{penedo2024fineweb,li2024dclm}. Domain reweighting has developed
along similar lines. Source partitions are being supplemented by semantic and
gradient-based ones~\citep{diao2025nemotronclimb,
DBLP:conf/emnlp/ShiZWFZZCZCZHXZ25}, and surrogates fitted on proxy runs,
reformulated objectives, and state-dependent policies now stand in for
full-scale mixture evaluation~\citep{DBLP:conf/iclr/LiuZMZDPJL25,
DBLP:conf/nips/Xie0DDLLLL0Y23,DBLP:conf/iclr/ChenHLCR25}. At both
granularities, the value of data is increasingly treated as dependent on the
model being trained.

\paragraph{Open questions.}
Three questions remain around transfer, decision granularity, and cost
reporting. For transfer across models, when do cheaper judgements carry over
to the target model? External scorers can be swayed by surface
style~\citep{klimaszewski2026cqfpitfalls}, and evidence that proxy models or
gradient-based partitions transfer across scales is so far only
empirical~\citep{DBLP:journals/corr/abs-2505-07293,
DBLP:conf/emnlp/ShiZWFZZCZCZHXZ25}. For decision granularity, what should the
unit of decision be? No domain definition is known to be generally preferable,
and although Proposition~\ref{prop:hierarchy} shows that the decision spaces
of selection and domain reweighting are not nested, the two are usually
applied in sequence, with
QuaDMix~\citep{DBLP:journals/corr/abs-2504-16511} as an early attempt to
couple them. Finally, for cost reporting, studies use different model scales,
token budgets, and hardware. Which accounting protocol can compare them while
separating amortisable setup from computation repeated for every target run?

\section{Data Strategies for Post-training}
\label{sec:posttraining}

Post-training adapts a pretrained LLM through SFT, preference optimisation,
and RL. The data used at this stage strongly affect the final model,
and a small, well-chosen subset can match or exceed training on the full
set~\citep{DBLP:conf/acl/DongYLLXL00ZZ24,DBLP:conf/naacl/LiZLCC0W0024}. Most
methods surveyed here target SFT; preference optimisation and RL are covered
less, and Section~\ref{sec:posttraining:summary} returns to the resulting open
questions.

\subsection{A Comparison of Methods}
\label{sec:posttraining:comparison}

Table~\ref{tab:posttraining:classification} compares post-training methods
along three axes exposed by the common formulation: the model-derived signal,
the outer objective or proxy, and the state queried by $\policy$. The
discussion below focuses on the first two axes.

\begin{table}[H]
\centering
\scriptsize
\setlength{\tabcolsep}{3pt}
\renewcommand{\arraystretch}{1.12}
\newcommand{\postgroup}[1]{%
  \rowcolor{rootcol!10}\multicolumn{3}{@{}l@{}}{%
  \textcolor{rootcol}{\textbf{#1}}}\\}
\tablestripes
\begin{tabular}{@{}L{0.30\columnwidth}L{0.38\columnwidth}
                    L{0.24\columnwidth}@{}}
\toprule
\tableheadrow
\textbf{Category} & \textbf{Signal or criterion} & \textbf{Example}\\
\midrule
\postgroup{Model-derived signal}
Distribution-based & Generated domain profile
& VersaTune\\
Gradient-based & Gradient direction or magnitude
& LESS, OASIS\\
\postgroup{Outer objective or proxy}
Average & Fixed-mixture target risk & LESS\\
Worst-group & Maximum group risk
& AutoMixAlign\\
Base-preserving & Target risk with drift control & VersaTune\\
Coverage proxy & Domain coverage
& Chameleon\\
\postgroup{State queried by $\policy$}
Data-only & No query of $s_t$ & Chameleon\\
Frozen-state & Fixed checkpoint $\theta_0$
& ADG\\
Evolving-state & Current $s_t$ and policy history & OASIS\\
\bottomrule
\end{tabular}
\caption{\textbf{Classification axes for post-training data strategies.}
Categories within each group are alternatives along one design axis.
Representative examples are shown, with citations in
Table~\ref{tab:posttraining:objective}. ADG denotes answer-divergence-guided
selection.}
\label{tab:posttraining:classification}
\end{table}

\keypoint{ptcol}{What model-derived signal does the data policy use?}
\noindent In Problem~\eqref{eq:data-policy}, $\policy$ may read the full training
state $s_t=(\theta_t,\xi_t,t)$. Most model-conditioned post-training methods
extract their value signal from the model component $\theta_t$, while some
stateful methods also use running statistics stored in $\xi_t$. The signal
matters because examples from a domain may provide less marginal benefit to a
checkpoint that is already strong in that domain than to one with weaker
coverage. Two recurring model-derived signals are distributional profiles and
gradients.

\begin{figure}[H]
\centering
\includegraphics[width=0.68\textwidth]{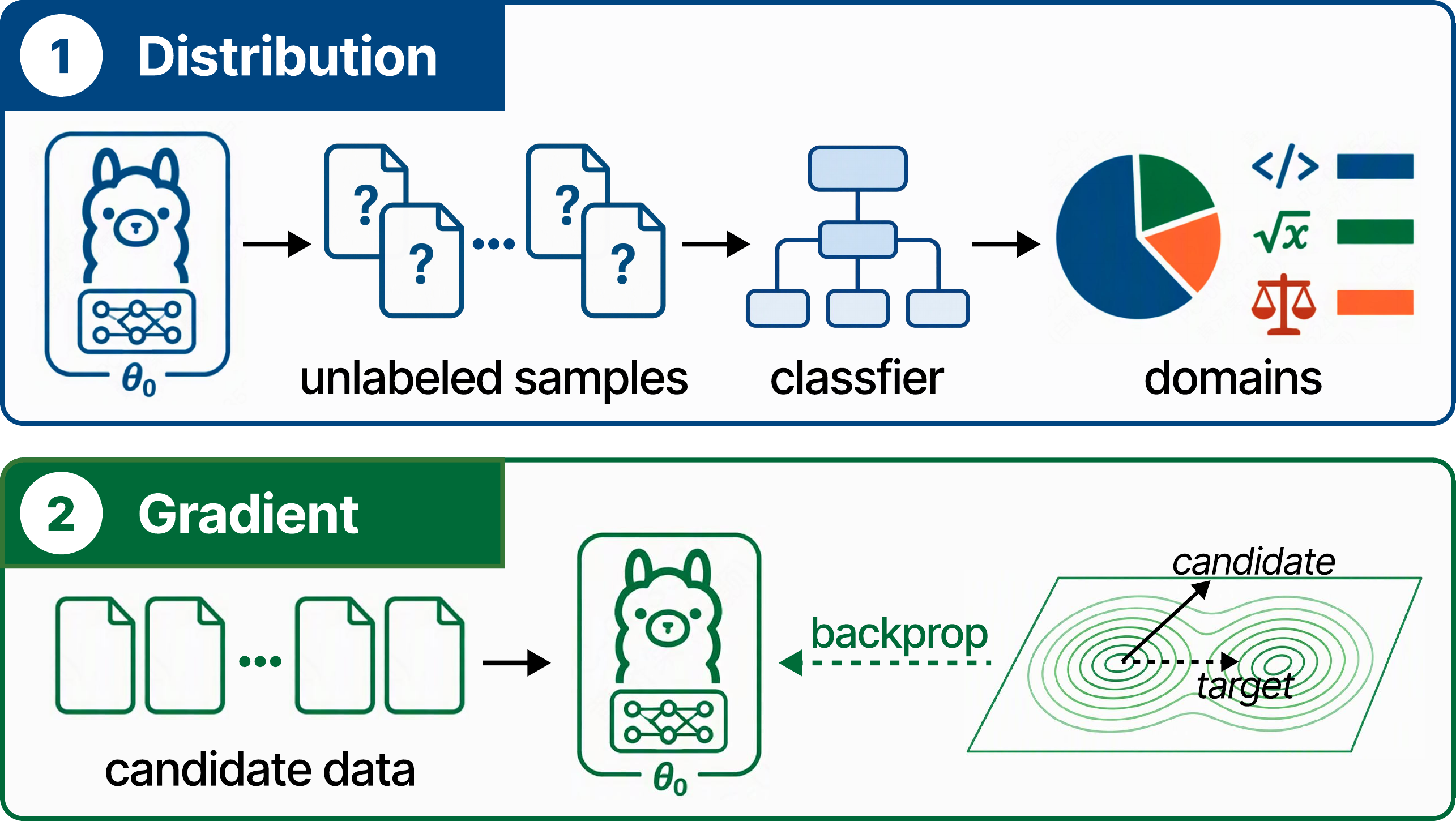}
\caption{\textbf{Extracting model-derived signals.}
Post-training policies query a model checkpoint to estimate data value from
generated domain distributions or local gradient information.}
\label{fig:model-state-characterisation}
\end{figure}

\begin{itemize}[leftmargin=1.2em,itemsep=2pt,topsep=3pt,parsep=0pt]
\item \textbf{Distribution-based characterisation} infers the pretraining
domain or knowledge distribution encoded by the model, typically by
classifying samples generated from it. VersaTune generates free-form samples
from the base model, classifies them into predefined domains, and averages the
domain probabilities to estimate the model's knowledge profile. This profile
initialises the SFT sampling proportions, which are subsequently adjusted
according to learning potential and forgetting
\citep{DBLP:conf/emnlp/LuZZLCWZ25}.
\item \textbf{Gradient-based characterisation} extracts the model's local
update geometry. LESS scores candidate gradients by cosine similarity to
target-task gradients~\citep{DBLP:conf/icml/XiaMGA024}, while EVIC extends
this signal to evolving inter-example interactions
\citep{DBLP:conf/icml/Liang00LWCH25}. OASIS
\citep{lee-etal-2026-oasis} uses the squared norm of each sample's last-layer
gradient as an approximate Fisher informativeness score, then standardises
that score against its exponential moving mean and variance. These running
moments belong to the auxiliary state $\xi_t$.
\end{itemize}

The two signals fail differently. Distributional profiles depend on a
predefined taxonomy and can miss variation within a domain. Gradient scores
are local to a checkpoint and target set, so their ranking can change during
training. Recomputing them tracks that change but adds recurring cost.
Neither signal directly estimates marginal utility after full training. A
coverage gap identifies where a model is weak but not whether the available
examples will close that gap. Gradient alignment is more local and can assign
high scores to examples whose update directions are already represented.
Cached rankings therefore require validation after the checkpoint or target
distribution changes.

\keypoint{ptcol}{How is the outer objective instantiated or proxied?}
\noindent Methods differ in the criterion through which $\policy$ is judged.
Most instantiate the outer objective $\objJ$ on the final model, whereas
methods without an explicit $\theta$-dependent objective use a proxy target to
construct $\policy$. Table~\ref{tab:posttraining:classification} summarises four recurring
choices, and Table~\ref{tab:posttraining:objective} lists
representative methods for each.

\noindent Let $\mathcal{E}_1,\ldots,\mathcal{E}_G$ be evaluation groups with
risks $L_g(\theta)=\mathbb{E}_{z\sim\mathcal{E}_g}[\ell(z;\theta)]$. Average-
and worst-group risk belong to the robust family
$\objJ_{\mathcal{R}}(\theta)=
\sup_{\rho\in\mathcal{R}}\sum_g\rho_gL_g(\theta)$, where
$\mathcal{R}\subseteq\Delta^{G-1}$. A singleton $\mathcal{R}$ gives a fixed
target mixture; the full simplex gives $\max_gL_g(\theta)$. AutoMixAlign
applies the latter to excess losses relative to task specialists. VersaTune
instead augments target risk with a penalty for forgetting non-target
domains. Coverage and instruction-following difficulty construct $\policy$
without evaluating $\theta_T$, so they are proxies rather than instances of
$\objJ$.
These objectives can rank the same candidate pool differently. Improving
average risk need not reduce the worst-group excess loss, while aggressive
target gains can increase forgetting. Evidence under one outer objective
therefore does not establish that a selector remains effective under another.
The proxy--objective distinction also limits what a reported gain supports. A
proxy-based result evaluates the proxy, policy, and training procedure
together; it does not show that the proxy is calibrated to $\objJ$.
Objective-based rankings can likewise depend on the composition of a small
target set. Cross-method comparisons should therefore hold the candidate
pool, training budget, and evaluation mixture fixed.

\begin{table}[H]
\centering
\small
\setlength{\tabcolsep}{5pt}
\renewcommand{\arraystretch}{1.16}
\newcommand{\objmark}{\textcolor{ptcol}{\cmark}}
\tablestripes
\begin{tabular}{@{}L{0.36\textwidth}C{0.12\textwidth}C{0.14\textwidth}
                    C{0.18\textwidth}C{0.10\textwidth}@{}}
\toprule
\tableheadrow
\textbf{Method} & \textbf{Average} & \textbf{Worst-group}
& \textbf{Base preservation} & \textbf{Proxy}\\
\midrule
LESS~\citep{DBLP:conf/icml/XiaMGA024}                 & \objmark & & & \\
DynamixSFT~\citep{DBLP:journals/corr/abs-2508-12116}  & \objmark & & & \\
OASIS~\citep{lee-etal-2026-oasis}                     & \objmark & & & \\
EVIC~\citep{DBLP:conf/icml/Liang00LWCH25}             & \objmark & & & \\
AutoMixAlign~\citep{DBLP:conf/acl/CorradoKDYZXPYC25}  & & \objmark & & \\
VersaTune~\citep{DBLP:conf/emnlp/LuZZLCWZ25}          & & & \objmark & \\
IFD~\citep{DBLP:conf/naacl/LiZLCC0W0024}              & & & & \objmark \\
ADG~\citep{li-etal-2026-instruction}                  & & & & \objmark \\
Chameleon~\citep{DBLP:conf/icml/XieTC25}              & & & & \objmark \\
\bottomrule
\end{tabular}
\caption{\textbf{Objectives and proxies for post-training data value.}
Average, worst-group, and base preservation denote their corresponding
training objectives; Proxy denotes a criterion that constructs the data policy
without evaluating the final model. IFD denotes instruction-following
difficulty.}
\label{tab:posttraining:objective}
\end{table}

\subsection{Summary and Discussion}
\label{sec:posttraining:summary}

\paragraph{Trend.}
Post-training policies range from data-only Chameleon and checkpoint-based
LESS and ADG to stateful OASIS, EVIC, and
DynamixSFT~\citep{DBLP:conf/icml/XieTC25,DBLP:conf/icml/XiaMGA024,
li-etal-2026-instruction,lee-etal-2026-oasis,
DBLP:conf/icml/Liang00LWCH25,DBLP:journals/corr/abs-2508-12116}.
Stateful policies can control inclusion, sampling frequency, and order; EDCO
spans all three~\citep{pang2026edcodynamiccurriculumorchestration}. Outer
objectives now include worst-group loss and preservation of base-model
capabilities~\citep{DBLP:conf/acl/CorradoKDYZXPYC25,
DBLP:conf/emnlp/LuZZLCWZ25}.
Greater adaptivity makes scores more specific to the current model, but also
shortens how long they remain valid. Comparisons that ignore rescoring
frequency can conflate a better signal with additional model queries.

\paragraph{Open questions.}
Fixed-pool supervised fine-tuning leaves three further questions. For
environment and task selection, where should an agent interact during RL?
Unlike fixed-pool SFT, environments differ in rollout cost, reward reliability,
and skill coverage. WebRL filters tasks generated from prior failures by
feasibility and critic-estimated difficulty, but only within WebArena. How to
allocate a rollout budget across heterogeneous environments remains largely
unexplored~\citep{qi2025webrl}. For selection efficiency, how can selection
remain cheaper than the rollouts it avoids? Curriculum RL with Off-Policy
Influence Guidance (CROPI) reuses initial-policy trajectories and compressed
gradients, yet recomputes features over a large candidate pool at each
phase~\citep{zhu2026cropi}. Partial pools, learned proxies, and reusable
rollout buffers reduce this work but may become stale as the policy changes.
Finally, for granularity and credit assignment, what should be selected in a
long-horizon task? Selecting demonstration steps and pruning state context
improves offline web-agent training~\citep{zadeh2026weasel}. Online RL jointly
chooses environments, tasks, trajectories, and transitions. Selection must
preserve trajectory coherence while attributing delayed terminal rewards to
intermediate decisions.

\section{Conclusion}
\label{sec:conclusion}

We cast selection, domain reweighting, and curricula as policies that allocate
a training budget over examples. Coarse decisions require a useful partition;
model awareness and adaptivity add scoring, proxy-training, or rollout cost.
Our analyses show that partitions depend on the construction signal and that
auxiliary floating-point operations (FLOPs) depend on model size, tokens per
run, and run count. Future evaluations should report utility and end-to-end
cost together, including for agentic RL policies.

\section*{Limitations}

This work focuses on text-only LLM data strategies. Selection, reweighting,
and curricula remain less studied for multimodal LLMs, where utility depends
on modality quality and cross-modal alignment. Analyses cover only text and
FLOPs.

\bibliography{sample}

\clearpage
\appendix
\begin{center}
{\large\bfseries Appendix Contents}
\end{center}
\vspace{2pt}

\begingroup
\small
\newcommand{\appentry}[1]{%
  \noindent
  \hyperref[#1]{\textbf{\ref*{#1}}\quad\textbf{\nameref*{#1}}}%
  \dotfill
  \hyperref[#1]{\pageref*{#1}}\par\vspace{2pt}}
\newcommand{\appsubentry}[1]{%
  \noindent\hangindent=3.2em\hangafter=1\hspace*{1em}%
  \hyperref[#1]{\ref*{#1}\quad\nameref*{#1}}%
  \dotfill
  \hyperref[#1]{\pageref*{#1}}\par\vspace{1pt}}

\appentry{app:related-surveys}
\appentry{sec:empirical}
\appsubentry{sec:empirical:setup}
\appsubentry{sec:empirical:rq1}
\appsubentry{sec:empirical:compute}
\appentry{app:notation}
\endgroup
\vspace{8pt}

\section{Related Surveys}
\label{app:related-surveys}
\label{app:survey-comparison}

\paragraph{General data selection.}
Prior surveys establish broad taxonomies for selecting training data.
\citet{albalak2024survey} organise language-model data
selection around a utility function and the mechanism that converts utility
into inclusion or sampling decisions, covering pretraining, instruction
tuning, alignment, in-context learning, and task-specific adaptation.
\citet{wang2023data} take a wider data-management
view encompassing quantity, quality, and domain/task composition, whereas
\citet{zhang2025survey} focus specifically on
quality-oriented subset construction for instruction tuning. We treat
selection, domain reweighting, and curricula as restrictions on a common
time-varying weighting policy. This framing supports comparisons across
training stage, decision granularity, state dependence, and selection compute
(Table~\ref{tab:survey-comparison}).

\begin{table}[H]
\centering
\small
\setlength{\tabcolsep}{6pt}
\renewcommand{\arraystretch}{1.15}
\tablestripes
\begin{tabular}{@{}L{0.42\textwidth}C{0.10\textwidth}C{0.10\textwidth}
                    C{0.10\textwidth}C{0.10\textwidth}@{}}
\toprule
\tableheadrow
\multirow{2}{*}{\textbf{Survey}} &
\multicolumn{4}{c}{\textbf{Training stages covered}}\\
\cmidrule(l){2-5}
\tableheadrow
& \textbf{PT} & \textbf{AN} & \textbf{SFT} & \textbf{PO}\\
\midrule
\citet{wang2023data}      & \cmark & \xmark & \cmark & \xmark \\
\citet{albalak2024survey} & \cmark & \xmark & \cmark & \cmark \\
\citet{qin2024tsunami}    & \xmark & \xmark & \cmark & \xmark \\
\citet{zhang2025survey}   & \xmark & \xmark & \cmark & \xmark \\
\midrule
\textbf{Ours}             & \cmark & \cmark & \cmark & \cmark \\
\bottomrule
\end{tabular}
\caption{\textbf{Training-stage coverage of related surveys.}
PT, AN, SFT, and PO denote stable pretraining, annealing, supervised
fine-tuning, and preference optimisation, respectively.}
\label{tab:survey-comparison}
\end{table}

\paragraph{Multimodal data selection.}
Data-centric MLLM surveys broaden the candidate unit beyond text to aligned
image--text pairs, audio, video, and 3D data.
\citet{bai2024surveymultimodaldata} follow the full multimodal data pipeline,
including collection, filtering and deduplication, pretraining selection and
modality mixing, adaptation, and dataset evaluation.
\citet{qin2024synergydatamultimodal} instead study data--model
co-development: multimodal data improve models, while models act as generators,
annotators, and evaluators of data. These surveys expose modality-specific
criteria such as cross-modal alignment and modality balance, but cover data
construction more broadly. We focus on the narrower optimisation question of
which existing training units to expose, at what weight, and at which training
state; under this formulation, multimodal methods change the utility signal and
group structure but share the same data policy abstraction.

\section{Empirical Study}
\label{sec:empirical}

\subsection{Experimental Setup}
\label{sec:empirical:setup}

We conduct two pretraining-focused analyses. The first compares source-,
semantic-, and gradient-based partitions of the same documents. The second
reconstructs auxiliary accelerator compute for representative
domain-reweighting methods under a shared target configuration.

\begin{figure}[!t]
  \centering
  \includegraphics[width=\textwidth]{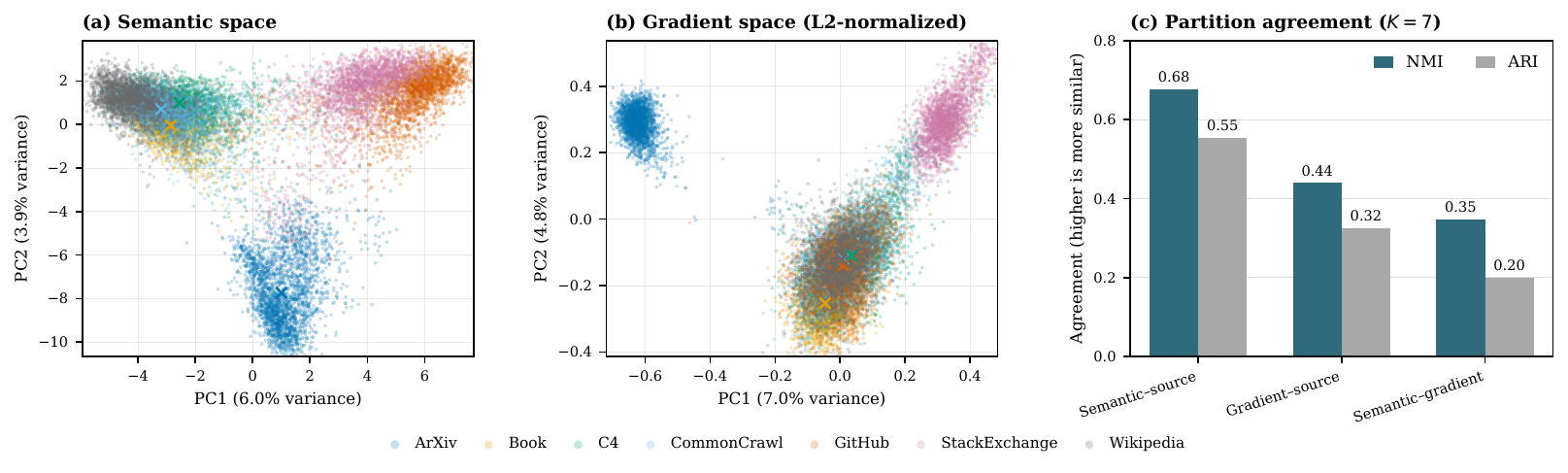}
  \caption{\textbf{Different signals induce different domain structures on the
  same SlimPajama sample.}
  (a--b) Principal component analysis (PCA) views of semantic and L2-normalised
  gradient representations, coloured by source. Crosses mark source medians. A
  stratified subset is shown for legibility, while clustering and agreement
  metrics use the full sample. (c) Agreement among source labels and the two
  induced partitions. Semantic clusters follow source boundaries more closely
  than gradient clusters, while semantic and gradient partitions have the
  lowest agreement.}
  \label{fig:domain-partitions}
\end{figure}

\subsection{RQ1: How Do Different Partition Signals Define Domains?}
\label{sec:empirical:rq1}

A domain is meaningful only relative to the signal used to construct it. We
therefore ask whether source metadata, semantic similarity, and model gradients
induce compatible partitions of the same documents.

\paragraph{Setup.}
We sample 115,839 SlimPajama documents, with at most 20,000 from each of its
seven source categories~\citep{cerebras2023slimpajama}. The source labels
provide the human-defined partition. For semantic-space domains, we encode
documents with
\texttt{stella\_en\_400M\_v5},\footnote{An English dense text-embedding model
released by NovaSearch for retrieval and semantic-similarity tasks. Here it
maps each document to a semantic vector for clustering.}
the encoder used by Nemotron-CLIMB
~\citep{diao2025nemotronclimb}. Gradient-space domains instead use projected,
L2-normalised gradients from Qwen2.5-0.5B-Instruct, following the TRAK
construction~\citep{pmlr-v202-park23c}. We cluster the first 50 principal
components with KMeans using ten initialisations and seed 42. The main
comparison uses $K=7$, matching the number of source categories, and reports
normalised mutual information (NMI) and adjusted Rand index (ARI).

\paragraph{Partition signals disagree.}
Semantic geometry preserves source-specific structure more strongly than
gradient geometry (Figure~\ref{fig:domain-partitions}). The two induced
partitions agree with each other even less than either agrees with the source
labels. The same ordering holds at $K=25$ and $K=100$, so it is not specific
to matching the number of source categories.

\finding{findingblue}{Source, semantic, and gradient signals induce
substantially different partitions of the same documents. A ``domain'' is
therefore a method-dependent modelling choice rather than a corpus-intrinsic
attribute.}

\subsection{RQ2: What Is the Auxiliary Training Cost?}
\label{sec:empirical:compute}

\begin{figure}[t]
  \centering
  \includegraphics[width=0.92\textwidth]{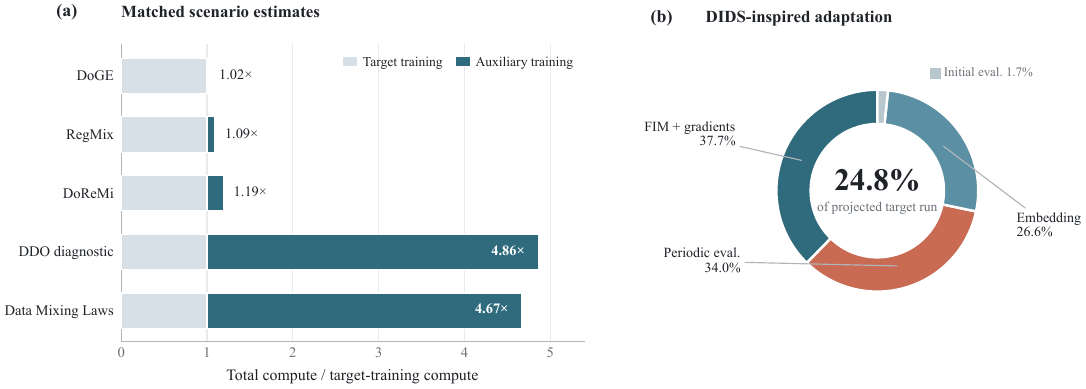}
  \caption{\textbf{Estimated auxiliary compute.}
  (a) Auxiliary-training FLOPs relative to a shared target run.
  (b) Logged component shares for a DIDS-inspired adaptation, using a
  projected target run as the denominator.}
  \label{fig:compute-cost}
\end{figure}

\begin{table}[t]
  \centering
  \small
  \setlength{\tabcolsep}{5pt}
  \renewcommand{\arraystretch}{1.12}
  \caption{\textbf{Auxiliary compute under a shared target configuration.}
  FLOPs use non-embedding parameter counts; method-specific tokens/run and
  overhead are measured against the shared target.}
  \label{tab:compute-cost}
  \tablestripes
  \begin{tabular}{@{}L{0.24\textwidth}R{0.13\textwidth}R{0.08\textwidth}
                      R{0.13\textwidth}R{0.20\textwidth}R{0.10\textwidth}@{}}
    \toprule
    \tableheadrow
    \textbf{Method} & \textbf{Proxy size} & \textbf{Runs}
      & \textbf{Tokens/run} & \textbf{Auxiliary FLOPs} & \textbf{Overhead} \\
    \midrule
    DoGE~\citep{DBLP:conf/icml/FanPJ24}
      & 42.5M & 1 & 2.911B & $7.423{\times}10^{17}$ & $2.05\%$ \\
    RegMix~\citep{DBLP:conf/iclr/LiuZMZDPJL25}
      & 1M & 512 & 1.050B & $3.226{\times}10^{18}$ & $8.92\%$ \\
    DoReMi~\citep{DBLP:conf/nips/Xie0DDLLLL0Y23}
      & 84.9M & 2 & 5.822B & $6.920{\times}10^{18}$ & $19.1\%$ \\
    DDO diagnostic~\citep{kang2024autoscale}
      & 1.035B & 30 & 0.3B / 1.2B & $1.397{\times}10^{20}$ & $386\%$ \\
    Data Mixing Laws~\citep{DBLP:conf/iclr/YeLSZZQ25}
      & 18.9--302.3M & 103 & 1.747B & $1.326{\times}10^{20}$ & $367\%$ \\
    \bottomrule
  \end{tabular}
\end{table}

\paragraph{Setup.}
The common target has 1.035B non-embedding parameters and trains on 5.822B
SlimPajama tokens. Following \citet{kaplan2020scaling}, a training pass costs
$6ND$ and a frozen forward pass costs $2ND$, where $N$ excludes token
embeddings. Adaptation is method-specific. DoGE and Data Mixing Laws preserve
their source proxy-to-target token ratios, DoReMi is matched to the target
budget, and RegMix retains its reported absolute budget. The DDO row retains
the 0.3B/1.2B scale-dependence diagnostic from AutoScale and is not a cost
estimate of its full extrapolation pipeline. Public run counts are retained.
The DoReMi estimate includes reference-model inference during proxy training.
CPU, I/O, and validation passes with unreported token counts are excluded, so
the table gives approximate lower bounds for a counterfactual common target.

\paragraph{Search structure determines auxiliary compute.}
Figure~\ref{fig:compute-cost}(a) shows that run count alone does not determine
cost. RegMix trains hundreds of ultra-small proxies, whereas the DDO diagnostic
uses fewer full-size runs. Data Mixing Laws combines many runs with several
proxy sizes. Model size, tokens per run, and run count must therefore be
reported together. This analysis does not measure mixture fidelity or
downstream utility, so it does not establish a quality--cost frontier.

\paragraph{Reuse changes amortised cost.}
Table~\ref{tab:compute-cost} reports auxiliary compute against one target run.
If the same proxy runs or fitted surrogate support $R$ target runs, their
amortised overhead becomes $C_{\mathrm{aux}}/(R C_{\mathrm{target}})$.
Checkpoint- or target-dependent scores may instead require fresh computation.
Cost reports should therefore state which artefacts are reusable and over how
many target runs. A change in the target distribution, corpus partition, or
scoring checkpoint can invalidate cached artefacts, so reuse should not be
assumed across settings.

\finding{findingblue}{Under a shared target denominator, estimated auxiliary
compute depends jointly on non-embedding model size, tokens per run, and run
count.}

\paragraph{A DIDS-inspired component profile.}
Figure~\ref{fig:compute-cost}(b) profiles a local adaptation based on
Qwen2.5-0.5B hidden-state embeddings and selection stages from reproduction
logs. Costs are normalised to a projected 80M-token target run. Published DIDS
instead clusters per-example gradients, so these component shares characterise
only the adaptation. Influence estimation and periodic evaluation account for
the largest shares, followed by embedding extraction.

\finding{findinggreen}{For this adaptation, evaluation frequency and reuse of
gradients or embeddings are the main efficiency variables to ablate.}

\clearpage
\onecolumn

\section{Notation}
\label{app:notation}

\captionof{table}{\textbf{Notation.} Symbols used in the common formulation
and its special cases.}
\label{tab:notation}

\begingroup
\footnotesize
\setlength{\tabcolsep}{4pt}
\renewcommand{\arraystretch}{1.22}
\newcommand{\ntype}[1]{\textcolor{rootcol}{\textsf{#1}}}

\noindent
\tablestripes
\begin{tabular}{@{}L{0.07\textwidth}L{0.105\textwidth}L{0.265\textwidth}
                  L{0.07\textwidth}L{0.105\textwidth}L{0.265\textwidth}@{}}
\toprule
\tableheadrow
\textbf{Symbol} & \textbf{Type} & \textbf{Meaning} &
\textbf{Symbol} & \textbf{Type} & \textbf{Meaning}\\
\midrule

\rowcolor{ptcol!8}
\multicolumn{6}{@{}l@{}}{\textcolor{ptcol}{\textbf{Data and training}}}\\
$t$ & \ntype{Index} & Current training step &
$T$ & \ntype{Scalar} & Total number of update steps\\
$\dataset_t$ & \ntype{Set} & Candidate pool at step $t$ &
$z_{t,i}$ & \ntype{Element} & Example $i$ in $\dataset_t$\\
$n_t$ & \ntype{Scalar} & Candidate-pool size &
$\mathcal{B}_t$ & \ntype{Multiset} & Sampled minibatch\\
$B$ & \ntype{Scalar} & Total token budget &
& & \\
\addlinespace[2pt]

\rowcolor{updcol!8}
\multicolumn{6}{@{}l@{}}{\textcolor{updcol}{\textbf{State and optimisation}}}\\
$s_t$ & \ntype{Tuple} & Complete training state &
$\theta_t$ & \ntype{Vector} & Model parameters\\
$\xi_t$ & \ntype{Variable} & Auxiliary optimiser or policy state &
$\ell(z;\theta)$ & \ntype{Function} & Per-example loss\\
$\widehat{\mathbf{g}}_t$ & \ntype{Vector} & Mean minibatch gradient &
$\update$ & \ntype{Function} & State-update map\\
\addlinespace[2pt]

\rowcolor{polcol!8}
\multicolumn{6}{@{}l@{}}{\textcolor{polcol}{\textbf{Data policy}}}\\
$\policy$ & \ntype{Function} & Data policy &
$\phi$ & \ntype{Parameter} & Policy specification\\
$\Phi$ & \ntype{Set} & Admissible policy class &
$\psi,\omega$ & \ntype{Parameters} & Curriculum parameters\\
$\Psi,\Omega$ & \ntype{Sets} & Curriculum hypothesis classes &
& & \\
\addlinespace[2pt]

\rowcolor{objcol!8}
\multicolumn{6}{@{}l@{}}{\textcolor{objcol}{\textbf{Allocation and objective}}}\\
$\mathbf{q}_t$ & \ntype{Distribution} & Allocation over the current pool &
$\mathcal{Q}_t$ & \ntype{Set} & Feasible allocations at step $t$\\
$\objJ$ & \ntype{Function} & Terminal target objective &
$\widehat{\objJ}$ & \ntype{Function} & Fitted surrogate for $\objJ$\\
$\mathcal{D}_{\mathrm{tgt}}$ & \ntype{Distribution} &
Target evaluation distribution &
$\Delta^{m-1}$ & \ntype{Set} & Probability simplex over $m$ units\\
\addlinespace[2pt]

\rowcolor{gencol!8}
\multicolumn{6}{@{}l@{}}{\textcolor{gencol}{\textbf{Structured decisions}}}\\
$\mathcal{P}$ & \ntype{Partition} & Corpus partition &
$\dataset_k$ & \ntype{Set} & Domain $k$ in $\mathcal P$\\
$K$ & \ntype{Scalar} & Number of domains &
$A$ & \ntype{Matrix} & Domain-to-example allocation\\
$\alpha$ & \ntype{Distribution} & Domain mixture &
$w$ & \ntype{Vector} & Generic mixture vector\\
$S$ & \ntype{Set} & Selected example indices &
$k$ & \ntype{Scalar} & Cardinality budget\\
$\mathcal{Q}_{\mathrm{rw}}$ & \ntype{Set} & Feasible example reweightings &
$\mathcal{Q}_{\mathrm{ds}}$ & \ntype{Set} & Feasible data selections\\
$\mathcal{Q}_{\mathrm{dr}}$ & \ntype{Set} & Feasible domain reweightings &
$r_\psi$ & \ntype{Function} & Curriculum signal\\
$h_\omega$ & \ntype{Function} & Scheduling rule &
$c_t$ & \ntype{Scalar} & Curriculum progression variable\\
\bottomrule
\end{tabular}
\endgroup

\end{document}